%% file: ddm-jdp-acs-July-2014.tex
\documentclass[11pt,letterpaper]{article}
\usepackage{cogsys}
\usepackage{cogsysapa}
\usepackage[T1]{fontenc}
\usepackage{times}
\usepackage[pdftex]{graphicx} 

\usepackage{color}

\usepackage{amssymb}
\usepackage{mathtools}
\input cup-macros.tex

\cogsysheading{3}{2014}{143--162}{9/2013}{7/2014}

\ShortHeadings{Representation of Inferences}
              {D.\ McDonald and J.\ Pustejovsky}

\begin{document} 

\title{Representing Inferences and their Lexicalization}
 
\author{David McDonald}{dmcdonald@sift.net}
\address{Smart Information Flow Technologies, 
        14 Brantwood Road, Arlington, MA 02476 USA}
\author{James Pustejovsky}{jamesp@cs.brandeis.edu}
\address{Department of Computer Science, Brandeis University,
         415 South Street, Waltham, MA 02453 USA}
\vskip 0.2in
 
 \vspace{-2mm}
\begin{abstract}
We have recently begun a project to develop a more effective and efficient way to
marshal inferences from background knowledge to facilitate deep natural language understanding. 
The meaning of a word is taken to be the entities, predications, presuppositions, and potential inferences that it adds to an ongoing situation.
As words compose, the minimal model in the situation evolves to limit and direct inference. 
At this point we have developed our computational architecture and implemented 
it on real text.
Our focus has been on proving the feasibility of our design.


\end{abstract}
\vspace{-2mm}

\vspace{-2mm}
\section{Introduction}  

Given the compactness of the lexicon relative to the number of objects
and relations referred to in the world,  ambiguity would seem to be 
inevitable. 
Compounding this problem is the fact that speakers
regularly omit information in what they say
yet 
their listeners fill it in without 
 conscious effort. In other
words, speakers leave \emph{gaps}, 
but somehow 
our semantic lexicon is  structured so as to fill in the holes in our
interpretation. 
This paper presents a  model 
for how this can be done. 

We begin this paper by laying out the problem we are addressing and our
assumptions. Section 2 will describe and motivate the new techniques and
representations we are using. Section 3 provides an example of how
they are used while  Section 4 covers the same ground in much greater
detail. We close with a section comparing our approach to others
and our plans for future research. 

\vspace{-2mm}
\subsection{Filling Gaps}
%
Semantic gaps are everywhere. 
Consider this text taken from a January 18th, 2006, Al Jazeera news article  about the first bird flu victim in Iraq:
\vspace{-2mm}
\begin{quote}
``{\it \ldots a 14-year-old girl died in the Kurdish city of Sulaimaniya \ldots The rest of the family is in good health \ldots}''
\end{quote}
\vspace{-2mm}

\noindent
We effortlessly know that this is 
the family of the girl, even across the three intervening sentences in the full text. The writer could  have said ``{\it the girl's family}'' but  did not have to, knowing that readers would supply this information through inference. 

Gaps like these are a pervasive and even essential component  of  language use: 
speakers appreciate what their listeners will infer 
from their knowledge of the world
(e.g., children are presumed to have families) and from the communicative context that they share.
It is one of the points of Grice's (1975) Maxim of Quality: 
do not be more informative than required. \nocite{grice1975logic}
%
The long-standing question, of course,  is how this is done.  How is our extensive body of background knowledge and inference organized? How do we deploy it so effortlessly?
That is the subject of this paper, 
where we lay out some of our initial results from a recently initiated project. 


\vspace{-3mm}

\subsection{Speed Implies Structure}  
\label{psychSpeed}
Psycholinguists have known for decades
that language comprehension is immediate, incremental,  
and works on all levels at once: syntactic, semantic, discourse, and pragmatic
\cite{marslen1973linguistic}.
People interpret utterances  
word by word without noticeable delay.
Recent work has shown that an event verb will activate its prototypical objects in just the time it takes to hear the verb and that this will influence the interpretation of later syntactic structures
\cite{matsuki2011event}.  

When cognitive psychologists explain this ability, they talk about people having \emph{schemas} that organize their knowledge of ordinary things and events 
\cite{BartlettSchemas}.
This resonates with the ideas and mechanisms of frames and scripts
that were developed in artificial intelligence and linguistics more
than thirty years ago \cite{minsky_frames_1975,fillmore76framesandnatureoflg}.
%
These mechanisms encode knowledge about conventional types of events and situations that people know about or have experienced: birthday parties, presidental inaugurations, 
eating at a restaurant, etc.  They provide expectations about what is likely to happen and what defaults to assume in order to account for things that must have happened but were not witnessed.

In areas of research such as neuroscience \cite{speer2009reading}  
or cognitive linguistics \cite{bergen2004simulated},
what a schema consists of or what it means, computationally, to `activate' a schema and `provide' expectations has different answers -- it is usually not the point of their research.
%
It is, however, the point of our own research.
This paper describes
our computational account of what schemas are,
how they are activated,
their mechanisms for controlling interpretation, 
and how they provide expectations, 
implicatures, and defaults.

\vspace{-3mm}
\subsection{The Importance of Knowledge}

%
The knowledge-rich approaches of the 1970s and 1980s were abandoned 
by main-stream natural language research as part of the move to `empirical' approaches that were made possible by the construction of large machine-readable text corpora and advances in machine learning \cite{church1993introduction}. 
At about the same time, a shift to ever-larger projects increased the salience of the ``knowledge acquisition problem'' -- that 
without a vast amount of knowledge, systems will be too brittle 
and will fail on anything outside of what has been expressly modeled. 
%
As a result, people working in natural language typically use 
shallow 
techniques that stop with just a description of what a text says and has none of the active, ``fill in the gap'' inferential capability that is critical for full, deep language understanding.

We agree that knowledge modeling is difficult. It is intellectually challenging to come up with
conceptualizations that have the requisite sensitivity to context, the capacity for composition, 
and associated expectations for actions and inference. 
But this background knowledge is absolutely needed if automated systems are to
learn from reading or
fully understand our instructions. 
%
We are not alone in this belief, as witnessed by the steady body of work by 
other groups  
\cite{LenSchubert2009,montazeri2011elaborating}. 
%
Moreover, there are now substantial knowledge stores 
to draw on. 
In addition to Schubert's KNEXT, 
there are 
ConceptNet  \cite{speer2008analogyspace}, 
FrameNet \cite{FStextUnder2001},
and the long-term products of the CYC project  \cite{guha1993cyc}.
Hence, 
we do not presume to do this by ourselves. 
Once our designs have been refined through testing on a realistic corpus against the series of prototypes we will implement, 
we intend to formalize our knowledge requirements and look for assistance from like-minded people in the language-centric part of
the knowledge-representation community 
for follow-on collaborations.

\vspace{-3mm}

\subsection{Our Research Focus}  



Our work focuses 
on how inferences are marshaled from
background knowledge when we use language.  
%
In order to focus our efforts, we have pushed to one side 
issues that we know are important parts of any operational solution, but which now would just be a distraction. 

\vskip 0.05in
\cbullet 
We are working from a corpus of written texts, not speech; 
\cbullet 
We are not dealing with dialogue; 
\cbullet 
We are not trying to acquire background knowledge automatically. 
\vskip 0.05in
\noindent
Instead, we are working out how highly efficient, lexically triggered inference and expectation can happen at all.
We are deliberately not yet invested in a particular ontology or a large knowledge store. We think it is more important to test and refine our computational machinery before drawing on the work listed above and working at a larger scale.

In the next section we lay out the elements of our architecture  
and summarize our claims. 
In Section \ref{sectionThree} we illustrate them 
with the example that we drew on when formulating our design. 
%
We follow this with a smaller, but thoroughly 
implemented, example in Section \ref{sec:suv-example} that we  walk through in detail.
We conclude with a discussion of related work and our future plans.

\vspace{-3mm}

\section{Representation: Situations, Predicates, and Packets}
\label{subsec:situation}

Every cognitive architecture has a notion of \emph{working memory}: some means of defining and delimiting what it will attend to and what it can be aware of at any given moment.
Every architecture also has a \emph{control structure}: a policy or
mechanism dictating what actions it will take and in what order. 

In our architecture\footnote{
The name C3 stands for ``the Compositional Construction of Context''.}
 -- \emph{C3} -- our working context is a 
\emph{situation}, 
where what we mean by `situation' is close to what it means
from situation semantics 
\cite{barwise1983situation}.  
We use a data-directed, event-driven control structure that
adapts techniques used in our language analysis engine 
\emph{Sparser} \cite{mcdonald1992efficient,mcdonald1996}.
%
%
We focus on the  notion of a ``situation type'': a reoccurring pattern of events and participants.\footnote{
The situation semantics literature has instead focused on situations
as a device that provides a denotation for a complex of events and
participants. } 
%
A populated situation accompanies an ongoing discourse 
and supplies the information that is latent in the words of a text.
In our view, situations hold the general world knowledge that perception unconsciously brings to mind.  They supply the bulk of 
information that lies below the perceivable tip of the iceberg.

At its base,
the situation holds representations of the entities, events, and predications 
that have been mentioned in the ongoing discourse. 
It provides a minimal model that consists of a set of typed structured objects.
For example, if the text is ``{\it a 14-year-old girl}''  then, when that phrase has been read, the situation contains representations of the girl, the age, and of the fact that the girl is described as being that age.

%

\vspace{-3mm}

\subsection{Lexicalized Pragmatics}
\label{packets}
In a lexicalized grammar, the terminals of the rules are specific words 
instead of lexical categories such as proper noun or transitive verb.
We propose to lexicalize meaning and inference 
-- to instantiate it 
directly from the incremental composition of 
the meaning of the words in a text 
without using an intervening logical form. 

The meaning of words, phrases, and meaning-bearing constructions is defined in terms of the set of 
entities, predicates, relations, propositions, or potential inferences they convey.
Situations are created dynamically by composing these \emph{packets} of content and inference as the words of a text are scanned.
Most packets correspond to small individual categories or inferences, such as
the affordances of a cup as a container or
the consequences of a process being canceled. 
Packets are small because they are designed to compose with other packets to collectively define the suite of inferences that are active in a situation.
Packets are activated singly or in groups according to what work they are designed to do and how and where they are triggered. 
The notion of {\it packet composition} is how we expect to satisfy one of the fundamental properties of language  recognized since the time of  von Humboldt:
 the ability to make infinite use of finite means. 




\vspace{-3mm}

\subsection{Predicates Linked to Language}
\label{black-predicate}

As a concrete example of a packet, 
consider the word {\it black}.
It is the English realization of 
the individual 
in the ontology that is used to represent the color black 
(denoted as {\tt\small\bf black}), 
as opposed to other colors such as red or titanium white. 
Like all colors, it is associated with a two-place predicate
that establishes a relationship between an entity that can have a color 
(tree leaves, cars, etc.)
and the specific color {\tt\small\bf  black}.
We encode this predicate as

\vspace{-2mm}

\begin{quote}
$\lambda x_{has-surface} [color\_of(x,black)]$
\end{quote}
 
 \vspace{-2mm}
 
 \noindent
where the type of object to which the predicate can apply is restricted:
it must include the type {\tt\small\bf has-surface}.
The object and the predicate together constitute the contents of the packet.
When the parser scans {\it black}, this packet is introduced into the situation.


Every predicate in
the ontology must specify what words or fixed phrases can express it
along with their linguistic properties.\footnote{
We use a Lexicalized Tree Adjoining Grammar for analysis and generation. A word's linguistic properties are established by its TAG tree family or families \cite{mcdonald1985tags,mcdonald1996}. }
The knowledge engineer adding colors to his conceptual model must indicate the word or phrase that names the color and that it has the syntactic patterns of a predicate adjective.
For C3, we do this using the notation for simultaneously defining semantic categories and their realizations described by McDonald (1994). \nocite{mcdonald1994reversible}



\vspace{-3mm}

\subsection{Latent Predicates} 
\label{subsec:latent-predicates}
When a phrase is fully instantiated, as in ``{\it a black SUV},''
the predicates receive values and establish predications. 
For example, the value of the color property of this SUV is bound to  {\tt\small\bf black}.
The meaning of substantive nouns or verbs will typically include a great many predicates, 
only a few of which will be present in a text and therefore explicitly represented as predications in the minimal model.
The other predicates are {\em latent}. 
They may be relevant as the text continues;
they may supply default assumptions that drive implicatures; 
or they may simply remain part of the background knowledge associated with the word, 
as we discuss in Section \ref{dismounting}.


In C3 we treat predicates formally as \emph{lambda variables}. 
These are structured objects defining a relationship between
individuals of a specific category,  
 constrained in the range of values they can take, i.e.,  what the variable can be bound to \cite{Krisp}. 
This information is self-contained within the object defining the variable: 
the category of individuals to which  it applies, the restrictions on  possible values, 
and the default values that can be assumed in the absence of actual ones. 

For example, 
if the  participants of an event are physical objects
then it is always the case that the event happened at some location,
even if we do not know its identity. 
When the analysis of our initial example had only gotten this far: ``{\it a 14-year-old girl died},'' we knew that the death must have happened at some location, but we didn't know what that location was.
The location could still be described, but only indirectly: 
``{\it where the girl died}'' or ``{\it the place where the girl died}.'' 
Once the text continued, ``{\it \ldots in the Kurdish city of Sulaimaniya},'' 
the latent variable that represented the location of the event is accessed and bound to that city.
Note that this narrows the category of the location to a {\tt\small\bf city}, 
and we would say ``{\it the city where the girl died}.''


In our implementation, 
a composite category  
defines all the possible properties, relationships, and habitats (see below)
that its instance individuals can have or can participate in, all represented by lambda variables.
\label{pre-anchored}
When we introduce a packet into the situation,  this potential becomes accessible, even when just a small part is present in the minimal situation model. 
%
We employ a wrapper around all variables, effectively a programming
trick, that lets C3  create an instance of each variable (potential
predication) linked to the relevant individual 
instantaneously in one step, at the moment the individual is introduced into the situation.

\vspace{-3mm}

\subsection{Frames and Habitats}
\label{subsechabitat-intro}
Packets are C3's building blocks. 
Most packets contain roughly the same amount of information as we
intuitively associate with a single word ({\it black}, {\it cancel}).
But of course there are relational structures that are considerably larger, 
structures that should be instantiated as a single unit 
but that have multiple parts and activities, 
such as an airport or a birthday party. 

For C3, we represent these as \emph{habitats}
\cite{pustejovskyGL2013}.  The notion of a habitat has its
intellectual roots in two places. The first is as an
extension and deepening of  
\emph{qualia theory} \cite{pustejovskyMIT1995}.  
We introduce a habitat into the situation all at once, but which
aspect of it is in focus (which gets priority in dictating
interpretations and making inferences) depends on what is in focus in
the text being read, as we illustrate in
Section \ref{qualia-as-scafolding}.  The term ``habitat'' deliberately
plays on the ecological metaphor to guide intuition as to what should
be included in a frame and what should not.

The other source for habitats is the knowledge representation
techniques of classical AI:
scripts for representing stereotypical events and episodic knowledge
\cite{SchankAbleson77}, 
and especially the notion of a frame \cite{minsky_frames_1975}. 
Minsky developed the concept of a frame during a seminar in the spring
of 1972 that was dedicated to Newell and Simon's (1972) book {\it Human Problem
  Solving},    
starting from Bartlett's (1932) notion of a
schema.  \nocite{BartlettSchemas} \nocite{NewellandSimonHPS}  
Over time, frames evolved into
today's RDF triple-stores and weakly expressive description logics, 
retaining just the notion of a taxomically organized classes 
as containers for ``slots'' (properties) that can be restricted to a
range of possible values.
%

We have returned to something close to Minsky's original conception, 
where  frame theory emphasized the transformations that would
occur as perspectives changed or scenarios progressed, 
and focused on frame recognition and repair to account for variations.
Inferences and other actions are tied to the creation of frames and to changes in their slot
values by invoking ``attached procedures.''  
Minsky's frame ``systems'' are mirrored in our habitats by 
sets of frames that are organized according the qualia they focus on
(see \ref{qualia}).
We are, however, using modern computational tools for abstraction and inheritance. 
Early knowledge-based language comprehension research used pre-build monolithic frames; 
ours are assembled dynamically according to what is actually needed
given the content of the text.

\vspace{-3mm}

\subsection{Indexical Functional Variables}
\label{subsec:indexical-variables}
The contents of a situation reside in a web of relationships and possibilities, 
most of them coming from the active habitats, others coming from the discourse relationships that structure the interpretation of the text, including relations that keep track of partial information as the text is being read.
To represent this, we use a set of indexical functional variables 
similar to those in the Pengi system \cite{agre1988thesis}. 
These variables designate constant, functionally identical relationships
within the processes of the system, while their values vary transparently to fit the moment-to-moment situation.

One of Agre's examples was the variable {\tt\small\bf the-cup-I-am-drinking-from}, 
which would be bound to whichever of the three cups that he kept in his office
that he was drinking from at the moment. The things he could do with this cup were always the same, 
while the identity of the cup would vary.
The actions the system takes are stated once in terms of indexical variables --
the presuppositions and significance of a functionally designated object is always the same.
Actions are not dependent on particular values,  
only on the function those values serve. 
Their actual values are managed automatically and transparently
according to the situation at hand.

\vspace{-3mm}

\subsection{Pegs}
\label{pegs}
In Pengi, the deictic variables are managed by its perceptual system.
In our framework they are managed by the parser
and identify the structure it has observed and the relationships it expects.
In most instances an indexical such as {\tt\small\bf theme} or {\tt\small\bf new} 
will be bound to specific individual, 
but since the situation is being updated incrementally as each word is scanned, there are always moments where a phrase is incomplete, its head and type not yet identified, but its impact on the situation still needs to be established.  
To do this we use Luperfoy's (1992) notion of a \emph{peg}. \nocite{luperfoy1992pegs}

For example, at the point in the parse where we have read just ``{\it a 14-year-old},'' 
the indexical variable {\tt\small\bf current-np-referent} 
is bound to a peg that was created when the parser scanned the ``{\it a}'' and recognized that it was starting a noun phrase that would have a referent.
The peg   
provides a place to accumulate predications 
and establish expectations. For example we know that
whatever this referent may turn out to be, it is something for which it makes sense to have an age measured in years.
The peg's properties are transferred to a regular individual once the head of the NP ({\it girl}) has been scanned.
Section \ref{two-people-peg} provides another example of this process.

It is an interesting psycholinguistic question whether earlier context
has already established the overall topic and narrowed the semantic
field from which the referent of an incomplete phrase like ``{\it 14-year-old}'' will be drawn. 
The term ``bird flu''  was in the title of the news article that this excerpt appeared in. 
Anyone familiar with the subject will know the types of
individuals that will be discussed and, given the age mentioned in the
phrase, will presume that it refers to a person. In other contexts, for example at a bar, the presumption might be that the 14-year-old was a single malt scotch. Whether people use such pre-established semantic fields or wait a moment to hear the head word is an open question that could be tested in a well-designed experiment.

\vspace{-3mm}

\subsection{Representational Principles and their Consequences}
\label{separate-axioms-from-language}
We have arrived at a set of principles for the representation of world knowledge in C3. 
These are an overlay on an otherwise conventional system of categories
and properties in a specialization lattice. The aim is to provide a flexible  link from language to the ontology
while retaining the economy of only having to state axioms and relation types once.
These principles include:

\vskip 0.05in
\cbullet Only add a category to the ontology if it makes a contribution, e.g., it adds predicates, state-change affordances, presuppositions, or defaults.

\cbullet No representation without realization. 
Every category should correspond to one or more words, phrases, features, or syntactic constructions.

\cbullet Predicates are only defined once; they may be restricted to different values at different levels in the category lattice but they retain their identity.
\vskip 0.05in

\noindent  In a conventional representation, there is a substantial distance in the specialization lattice between the particulars that appear in a text,
such as a sport utility vehicle, which will be close to the bottom,
and what we know about the vehicle,  e.g.\ that it is a  {\tt\small\bf container}, which is stated at a high level and applies to a great many things besides SUVs.
It is difficult to use language in such a system.
Our need to have packets for domain-specific words that refer to 
general predicates and affordances (our \emph{lexicalized pragmatics}) cannot be easily accommodated.

\vspace{-3mm}

\subsubsection{Unique variables}
We chose instead to separate the realization facts (what words and construction are used)
from the axiomatic facts (what predicates and operations apply and what follows from them).
In C3, an SUV acts like a container because its category literally incorporates the {\tt\small\bf container} category
and uses its variables
to express the affordances available to its passengers and to state facts such as when one passenger gets out there is one fewer inside. 

We do this by making all variables (predicates) unique. They are defined once, as one object in the representation, on a category as far up in the lattice as possible for maximal application.
On more specific categories a variable will usually be restricted.
For example the {\tt\small\bf contents} variable of {\tt\small\bf container} is defined there as a collection of an unknown number of entities of unknown types.
When we move down to, say, {\tt\small\bf passenger-transporter} 
(see Section \ref{passenger-transporter}),
the type of the collection is restricted to {\tt\small\bf person}.
On a particular type of {\tt\small\bf passenger-transporter}, say {\tt\small\bf airplane}, the restriction on the variable will be further restricted to the different roles of people on an airplane.

The vocabulary is stated against these restrictions. 
Any packet that includes  {\tt\small\bf container} adds to the situation model the fact that its
contents are in one of two states, expressible as being {\it in} ({\it inside}) or {\it out} ({\it outside}) of the container, and have the affordance of being able to move between these states.
But we say that we {\it take} or {\it pick out} jelly beans from a jar (they cannot move on their own).
We watch a squirrel {\it climb out of} a garbage can (they can move on their own, and the movement involves ascending a height). 
When the variable is restricted to the category {\tt\small\bf person}, we refer to    
 {\it passengers} or use their roles ({\it driver}, {\it pilot}, {\it steward}), and they {\it go into} or {\it get out of} the container.

\vspace{-3mm}

\subsubsection{Pre-cached, ``Composite'' Categories}
\label{composite-categories}
Allowing different local restrictions on the same predicate object
lets us achieve an economy of expression for axioms, which is essential for working with large ontologies,
while retaining flexibility in how to define packets of the vocabulary 
since realization facts can refer to restriction categories at very different levels in the ontology.
But this comes at a cost, since any word with a rich meaning will have a packet that introduces dozens if not hundreds of latent variables (particularly for habitats) that will entail including a proportional number of categories.

We make this manageable by using what we call \emph{composite categories}. 
We define them as a conjunction of regular categories.
We then pre-cache the categories' variables (with their restrictions) to create a single computation object.
The result has the behavior we would get by using ordinary inheritance, but with none of the costs of traversing the lattice to collect the variables and apply their restrictions.
%

While a composite category often just collects the categories that are above it in the hierarchy, 
there is no requirement that it do so. Categories from very different parts of the ontology can be incorporated into a single composite. This makes for an ontology that is easier to maintain, since there is no requirement to force everything into a single lattice with single lines of inheritance.
Composite categories can be incorporated into other composites. When this happens, the incorporated composites are treated like macros that are unpacked inline and repackaged as a new class.\footnote{
 We work in Lisp and make heavy use of the multiple inheritance capabilities of the Common Lisp Object System \cite{gabriel1991clos}.}

\begin{figure}[t]
\vskip 0.05in
\begin{center}
\centerline{\includegraphics[width=0.8\columnwidth,clip]{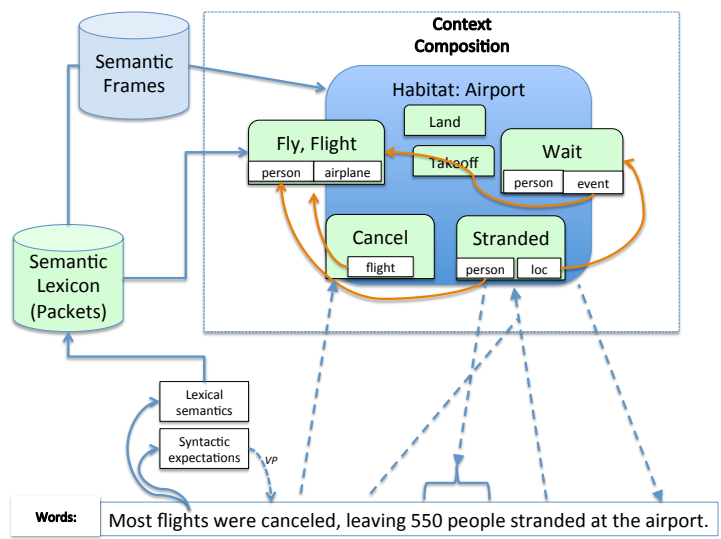}}
\vspace{-15mm}
\caption{The C3 architecture. As described in Section \ref{sectionThree}, the
  airport habitat includes a latent representation of its normal
  entities, roles, and activities. The C3 analysis incrementally
  brings some of these into focus, instantiating relationships and
  grounding otherwise anonymous text references such as the 550 people.}
\label{fig:arch}
\end{center}
\vskip -0.2in
\vspace{-5mm}
\end{figure}


\subsection{The C3 Architecture}

Figure~\ref{fig:arch} shows the basic framework of C3. 
Solid blue lines from the text trace the activation path up from the first part of the text to add packets (in green) or larger habitat frames (in blue) 
to the situation as a whole (outer box).  Dotted lines show later additions to the situation (upward arrows) or inferred interpretations made by the situation (downward arrows).  Orange arrows within the situation sketch relationships developed among the packets by binding variables.


C3's workflow begins with the perceived input; in our research this is the sequence of words in a text.  Words are interpreted as they are reached by the parser and contribute packets of content of different sizes and function to a growing situation.  This leads to the instantiation and assembly of highly structured sets of prototype relations and events, anticipated scenarios, and specific or prototypical individuals, places, and the like.  The situation then governs the expectations and interpretations of words and phrases as the analysis continues.


The C3  architecture assumes that utterances are
interpreted incrementally, making use of inferential packets which
drive the compositional construction of meaning. The result of the
interpretation process is a minimal  simulation of the situation
denoted by the utterance.

\vspace{-3mm}

\subsection{Our Claims}

We make two principal claims about the nature of natural language understanding  
as a computational process: 

\vskip 0.05in

\cbullet Language understanding is an incremental process,  
where all levels of analysis -- syntactic, semantic, and pragmatic -- 
are carried out simultaneously. 

\cbullet This process is governed by a highly structured, predictive model of the
ongoing situation 
that actively incorporates our background knowledge of the world. 

\vskip 0.05in

\noindent{}These claims are consistent with what is known 
from psycholinguistics 
about the language comprehension process in people 
-- the only example we have of fully effective language processors 
(see Section \ref{psychSpeed}). 
However, substantiating these claims from 
a computational perspective 
requires an implementation to establish that the claims are coherent
and to provide a platform for experimenting with different mechanisms
and representations. 
The two sections that follow illustrate 
how different facets of the new or newly revived technical capabilities we have just
described are deployed. 
In particular: 

\vskip 0.05in

\cbullet Organizing the meaning of words as `packets' 
of model-level content along with overt
and implicit predications  (in Section \ref{ambiguous-words}).

\cbullet Providing a partial, incremental predictive representation of 
phrases' referents as they are read (in section \ref{two-people-peg}).

\cbullet Using the situation to provide defaults through coercion
and  as a means of
creating a real-time simulation of an utterance as it unfolds 
through the semantic interpretation process  
(in Section \ref{stranded-passengers}).

\cbullet Treating situations computationally 
as the sum of the understanding of what has been said, along with what is
implied and what might follow (in Section \ref{compositionality}). 


\cbullet  Providing  functional landmarks to the content of a
situation to permit one-step application of anaphoric-style
inferential gaps (in Sections \ref{initialSituation} and \ref{dismounting}).

\vskip 0.05in

%

%

\vspace{-3mm}

\section{Creating and Applying a Situation}  
\label{sectionThree}
In this section we describe how the situation is established and drives inferences 
during C3's comprehension of an utterance. 
We focus on this text. 

\vspace{-.2cm}
\begin{quote}
``{\it Most flights from the Luis Munoz Marin Airport in San Juan to the Leeward Islands were canceled Monday, leaving about 550 people stranded at the airport.}'' \footnote{
This is a self-contained excerpt from a news article about the impact of Hurricane Earl on Puerto Rico (The New York Times, August 31, 2011).}
\end{quote}
\vspace{-.2cm}

%
\noindent
Taken just for its literal content, 
as most of today's language understanding systems would do, 
%
the result leaves many questions open. In particular, 
where did these 550 people come from, and why are they stranded?
In the section below, we show how lexical semantic knowledge
associated with the words in this example direct our inferences
towards ``filling in the gaps'' in the literal assertions from the text. 
We then demonstrate how packets of information are formed from lexical
items and how they compositionally build contextually salient inferences.




\vspace{-3mm}

\subsection{Lexical Structures}

Outside of a specific context, most high frequency words are ambiguous. Even once a word sense has been determined, there are still differences in logical perspective to sort out or metonymies to decode.
We describe our approaches to these problems in this section.

\vspace{-3mm}

\subsubsection{Simple ambiguity.}
\label{ambiguous-words}
Consider the word {\it flights}, 
which has different meanings in different domains. It could refer to a {\it flight of stairs} or be part of a fixed phrase like {\it flight from stocks}. It could refer to a quantity of beer or champagne or it could be a nominalization of {\it flee}. A fully populated language understanding system would have all of those readings and more. 
In the context of this example it refers to an airline flight,
but C3 must establish that fact 
before it can instantiate the {\tt\small\bf air-travel} habitat and activate its affordances.

We know from psycholinguistic studies that all of the senses of a
polysemous word are available for about 250 msec after the word is
read 
and that after about 500 msec, roughly when the next word has been read,
only the contextually appropriate sense is available 
\cite{swinney1976effects}. 
The context provided by the situation is sufficient for people to rapidly and unconsciously
disambiguate words that are ambiguous in
isolation like {\it flight}. The question is how does this happen. 
 Sublexical' techniques have been explored, including marker
passing \cite{charniak1983passing} and lateral inhibition
\cite{cottrell1983connectionist} though only in small systems. 

In C3, each kind of `flight' that it knows about (for which is has a
packet in its lexicon) has its own projection to the grammar, and will
introduce its own semantically-labeled reading into the analysis when
it is scanned, such as {\tt\small\bf airline-flight} and {\tt\small\bf flight-amount}. 
This mirrors the observed immediate activation of all the word's
senses. 
When the next word is scanned, the word {\it from} in this example, it
introduces its own projections, including its possibilities for
composition in  C3's lexicalized semantic grammar. 
This lets us use a simple disambiguation policy: only senses that
can extend through composition with the phrases around them can have their
meaning incorporated into the situation. The others are ignored. 
In this example we get {\tt\small\bf airline-flight} because the
preposition {\it from}  is part of the rule pattern that applies to
`flights' as movement 
(i.e., ``flights \emph{from} the Luis Munoz Marin Airport'').

As we suggested earlier, another possibility is that in an ongoing,
established context such as news about a hurricane, the set of available
readings for ambiguous words has already been narrowed to just those
that are applicable in that semantic field. The psycholinguistic
studies of lexical access \cite{lexialAmbiguityCollection1988}
may well be based on stimulus conditions and probes that do not apply
in the normal use of language between interlocutors who know they are
in a shared situation. This would replace the problem of word sense
disambiguation with the more realistic problem of recognizing the
situation type.
We intend to investigate this question in our future work.



\vspace{-3mm}

\subsubsection{Lexical entries in the generative lexicon.}
\label{GL}
In Pustejovsky's (1995, 2013b) Generative Lexicon theory,
\nocite{pustejovskyMIT1995, pustejovsky2013type}
the lexical entry for a content word 
(as opposed to a grammatical function word such as {\it most} or {\it from}) 
encodes three kinds of information: 
\vskip 0.05in
\cbullet Its {\bf argument structure}, which spells out what arguments the word takes, how they are realized syntactically and govern semantic role selection;

\cbullet Its {\bf event structure}, its class of event
(state, process, transition) and how it structures its 
implicatures \cite{pustejovsky1991syntax};

\cbullet Its {\bf qualia structure}, the basis of logical polysemy, implicated in coercion and type shifting. 
\vskip 0.05in
\noindent
The argument structure is integrated into the rule sets of the grammar 
and helps with simple disambiguation. 
The event structure is part of the habitats that are added to the situation 
and provides a scaffolding for anchoring events and action sequences. 
The qualia structure organizes the applicable predicates and affordances.

\paragraph{Qualia and logical polysemy.}

The Qualia consist of four basic roles, each of which can be seen as answering  a specific question about its associated object. Each contributes a complementary set of latent predicates to a word's meaning:

\label{qualia} 
\vskip 0.05in

\cbullet {\bf Formal roles} encode taxonomic information about the lexical item (the {\tt is-a} relation).
{\em What kind of thing is it; what is its nature?}

\cbullet {\bf Constitutive roles} encode information about the parts and constitution of an object ({\tt\small\bf part-of} or  {\tt\small\bf made-of} relation).
{\em What is it made of;  what are its constituents?}

\cbullet {\bf Telic roles} encode information on purpose and function (the {\tt\small\bf used-for} or {\tt\small\bf functions-as} relation).
{\em What is it for; how does it function?}

\cbullet {\bf Agentive roles} encode information about the origin of the object (the {\tt\small\bf created-by} relation).
{\em How did it come into being; what brought it about?}

\vskip 0.05in
\noindent
Most words have alternative readings that are characterized by
different qualia: the newspaper you read (telic), the one you spill
coffee on (constitutive), the one whose editorial opinions you
disagree with (agentive).  This distinction is referred to as {\em
  logical polysemy} \cite{pustejovsky1993lexical}. 
 Once a content word has been narrowed to the domain where it has a specific meaning (simple disambiguation), the next step is to determine its qualia role, to disambiguate it logically.

The qualia role that applies in a particular instance cannot be determined independently of the rest of the context.  If the text was {\it My flight just landed}, it would be the constitutive role, since we are talking about the airplane that the flight used and only physical things can land.  If our flight was rescheduled, it would be the agentive role.
All of these alternatives are part of the {\tt\small\bf air-travel} habitat -- a frame that 
factors into different parts (incorporated habitats) according to which qualia is involved. 
In this instance of {\it flight},\footnote{
Recall that the context is {\it ``Most flights from the Luis Munoz Marin Airport in San Juan to the Leeward Islands were canceled Monday \ldots'' } }
it is the telic role and it links to 
the portion of the habitat  
that organizes knowledge about flights as conveying people 
from place to place. 
\label{qualia-as-scafolding}

\vspace{-3mm}

\subsection{Habitats, Actions, and Composition}
\label{sec:air-travel}

Airports have control towers, runways, taxiways, gates, and
terminals. These are all available in the airport habitat.  
These are entities and relationships that the habitat knows about, 
but they are latent rather than part of the situation's minimal model. 
The principal activity at airports is air travel, and, 
if we ignore its personal aspects (making reservations, getting to/from the
airport, buying food, shopping), 
the most salient aspect of air travel is the flights.  
Flights are also habitats. 
They have a plane (the equipment), a crew, passengers, baggage, and food. 
They are run by particular airlines, have a flight number, and travel from one airport to another.


\label{passenger-transporter}
 In the telic reading of {\it flight}, the habitat includes a script that lays out the typical sequence of events and activities that constitute air travel.  Airplanes are containers and they can move. Like any moving container, when they move (taxi, take off, fly, land), they convey their contents with them from their starting point to their destination.
%
%
There are enough of these `passenger-transporters' in the world that they form a useful composite class: cars, buses, trains, bicycle-pulled carts, trucks, and others. This ensures that their common core is shared, particularly, for our purposes, the words that accrue to this level, such as {\it passenger}.

The interpretation of {\it flight} is as a process.  There is a state of affairs that holds before this process starts and a different one after it ends.  The principal difference between these two is in the location of the airplane and its contents: the passengers, their baggage, the crew.  Before the flight leaves they are at the origin airport, afterwards they are at the destination airport.
Any habitat like {\tt\small\bf flight} that involves scheduled process comes with the default assumption that once the process has started it will continue until it ends.

To represent the content of the first part of this text, 
C3 instantiates a flight habitat with values for the variables that we know. 
This adds to the situation a collection of an indefinite number
of individual flights, where each of these otherwise unidentified
flights originates in San Juan and terminates in an airport in the
Leeward Islands.  Each of these flights has a carrier and a flight
number, a crew and a passenger manifest, but these are latent
properties whose values are unknown, just as we do not know the actual number of flights in the collection.


\paragraph{Compositionality} 
\label{compositionality}

{\it Cancel} is an operator over processes: it modifies the situation rather than simply adding to it.  
Its syntactic configuration (as main verb) establishes that it 
applies to 
the value of the functional variable {\tt\small\bf syntactic-subject}, i.e., the flights.
Since the only qualia of flight that involves a process is its telic function of transporting its passengers from one place to another, that aspect of the {\tt\small\bf flight} habitat becomes central to the situation. 

Applying the operator {\tt\small\bf cancel} to the flights cancels this process. 
To cancel a flight means that it does not start (the flight does not {\it take off}).
This modifies the situation to reflect the fact that the conditions that held before the process would have started still obtain: the passengers who would have been on the flights are still at the San Juan airport, as are the crews and the planes.

\paragraph{Situation-driven binding.}
\label{stranded-passengers}
In the last portion of the canceled flights example, we have a result
clause, 
``{\it leaving about 550 people stranded at the airport}''.
Given its form, the syntactic relation of this adjunct to its main clause tells us that this state of affairs (the stranding of the people) happened because of the event in the main clause (the cancelation of most of the flights).  Being stranded is a habitat in itself, associated with air travel but not a part of it per se in the way as, say, losing one's luggage.
The meaning of {\it stranded} is that there was an intention to move that has been blocked: the path of the passengers' expected futures has been interrupted.
Note that the airport employees are not stranded, because they have a different
role in the {\tt\small\bf air-travel} habitat, i.e., they work at the airport. 

Inferences should be guided by what is salient in what is perceived -- the text that C3 is interpreting and the situation model created for it.  The cancelation brings into focus within the situation those elements that were most affected by it: the passengers, the air crews, and any other individuals whose intended future path of events was shifted.
This salience makes it simple to interpret the two definite references in the result clause.  
Given the context provided by this situation, we can bind the referent of {\it the airport} to San Juan's Luis Munoz Marin airport because the {\tt\small\bf flight} habitat has already created properties for two airports (origin and destination).  The origin airport is the more salient of the two because it is the one impacted by the cancelation.
Similarly, the {\it 550 people} are resolved to be the only people who are made salient 
by the cancelation: the passengers and crew who would have been on the flights 
that did not take off -- did not follow their intended, default future path.

This section has illustrated our claim that language understanding is
governed by 
knowledge-rich, predictive models of the ongoing situation
We have shown how this makes it simple to draw complex inferences in C3.  
We first recognize and instantiate the appropriate situation type (``activity at an
airport'').  That large habitat is focused on a particular qualia as
the text is incrementally interpreted (``most flights''), and
specialized through composition as C3 continues reading and
introducing packets into the situation (``canceled'').  This provides
the context in which the identity of the ``550 people'' is immediately
established, because they have the situation's role of {\tt\small\bf  passengers}, made salient by the cancelation of their flights.
In the next section we will walk through this process in detail on
a smaller, fully implemented example.

\vspace{-3mm}

\section{A Detailed Example: ISR}  
\label{sec:suv-example}

We have access to a set of logs of actual text-chat collected from an
Intelligence, Surveillance, and Reconnaissance (ISR)  team during the Empire Challenge 2010 
military exercise.
These are from a 
team
that was composed of three camera operators, an analyst, and a coordinator, 
all communicating over Internet Relay Chat, 
reporting on the movements and activities of other players in this live Army exercise in a simulated set of Afghani villages. 
This excerpt illustrates the sort of gap that we are focusing on. 
Camera operator Heavy2 is reporting on an event involving a car `of interest' in the Wakil village that he is observing.



\begin{table}[t]  

\vskip -0.25in

\begin{small}
\begin{center}

\caption{Team chat excerpt from Empire Challenge 2010}
\vspace {5mm}
\label{table:chat}

\vskip -0.10in
\begin{tabular}{|l|c|c|l||}  

\hline
Line &Time &Message \\ 
\hline

72 & [19:51] & \multicolumn{1}{|l |}{\(<\)Heavy2\(>\) black ford suv has entered wakil} \\
\hline
73 & [19:52] &  \multicolumn{1}{|l |}{\(<\)Heavy2\(>\) two people are dismounting} \\ 
\hline

\end{tabular}
\end{center}
\end{small}

\vspace {-5mm}

\end{table}

\noindent
It is obvious to us where the people came from. 
In this section  we lay out how we make it equally obvious to the C3 System. 

\vspace{-3mm}

\subsection{The Initial Situation}
\label{initialSituation}
Line 72 of the chat transcript, entered at 19:51 pm, is the first time that observer Heavy2 has typed anything for several minutes. This speaker shift has cleared the situation of any active habitats or facts, and moved their content to a passive store from which they can be reactivated when mentioned again. 
%
In this case, the
``{\it black Ford SUV}'' 
was already identified and designated as a `vehicle of interest' earlier at 18:27, and at 18:50 there was the report ``{\it three guys have gotten in to black ford suv at wakil}.'' Not only is there a known individual to add to 
the situation
(rather than building a new individual), but something is already known about it:\footnote{
The expressions used in this section are purely notional for purposes of illustration.
In C3's implementation their equivalents are configurations of typed objects linked by pointers and organized by indexical-variables bound by the situation object. 
We cannot describe their actual elements and organization in the space available.
}

\vspace{-3mm}
\begin{quote}
{\tt\small
SUV-1: container.contents = collection(count > 3, type = person).
}
\end{quote}
\vspace{-3mm}

\noindent{}The discourse history established that
the SUV is value of the {\tt\small\bf given} indexical variable.
The value of the {\tt\small\bf new} variable is the fact that it has entered the village.
This reintroduces the already-known village into the situation model, along with the fact of the event, but nothing else. 
The present location of the SUV is known (it is part of the minimal model),
but nothing is known about its previous location except that it had one: 
\label{prior-location}
``{\it where the SUV was before it entered Wakil}.''

From the text there is nothing else known
about the SUV, not even whether it has stopped moving.
But in the actual world of the observer,
all of this is an established part of reality: 
It approached along a particular road at a particular angle to the viewer;
the sun was shining, creating a shadow of a particular size; 
 buildings in Wakil are made of concrete and painted some color.
All of this is true, but only what is actually given in the text is present in the situation. The rest is latent.

\vspace{-3mm}

 \subsection{Expectations}
\label{two-people-peg}
In C3, texts are parsed incrementally word by word 
so as to get the greatest amount of leverage 
from the situation. 
From line 73, reported a minute after the report about the SUV,
C3 reads the word {\it two}.
As a nominal premodifier, this deploys a peg and its packet establishes that there is a collection of size two, but that is all that is known at that moment.
The rest of the text could refer to 
two of the windows on the SUV being opened, or two of its doors:

\vspace{-3mm}

\begin{quote}
{\tt\small
Peg(x): collection(count = 2, type = x)
}
\end{quote}

\vspace{-3mm}

\noindent
Upon reading {\it people}, the head of the NP, the peg is replaced by an individual representing a collection of two people, but again we know nothing more. 
%
There is an expectation, however.
The people must have been somewhere before this, even if we do not yet know where. 
Since some things, like the locations of the objects of discourse, are essential to understanding
(physical objects do not just appear in a puff of smoke).
This information gap leads to an expectation
that we will either be told the location or should assume one given the available evidence:

\vspace{-3mm}
\begin{quote}
{\tt\small
people-2: type = collection-2,  physical-object.location = ?).
}
\end{quote}
\vspace{-3mm}

\vspace{-3mm}
\subsection{Composition}
\label{dismounting}
C3 then reads the verb group of line 73, ``{\it are dismounting}.'' It adds the packet for {\it dismount} to the situation and
notes that this is an ongoing action:

\vspace{-3mm}

{\tt\small
\begin{quote}
\begin{verbatim}
dismount = transition.inprogress 
  movement.from = high 
  movement.to = low(ground)
  movement.actor = v:subject
\end{verbatim}
\end{quote}
}
\vspace{-2mm}

\noindent
From the syntactic construction,
it knows that the collection of people supplies the obligatory argument to {\it dismount}: who is doing the action.

{\tt\small\bf Dismount} is a movement. 
Every instance of a movement comes with predicates for where its participants 
(the two people who are moving) 
were before the action and where they are after it. 
None of these values have been given explicitly, although a firm default for {\tt\small\bf dismount} is that the final location is the ground. (One dismounts from a horse or a piece of gymnastics equipment.)

To establish the value of their prior location (where they dismounted {\it from}), C3 uses what amounts to anaphoric reasoning: namely,  what are the known locations given the present situation? This gives us the village and the SUV, but the SUV should be preferred because the thing one dismounts from must be close by
(compare ``{\it two people are walking up to it}'')
and the SUV is salient because it is the value of the discourse {\tt\small\bf theme} indexical because it is a `vehicle of interest':

\vspace{-2mm}

{\tt\small
\begin{quote}
\begin{verbatim}
during.before(dismount-1): 
  people-2.physical-object.location = SUV-1
  dismount-1.movement.from = SUV-1
\end{verbatim}
\end{quote}
}
\vspace{-2mm}

\noindent
This binding has significant side effects. 
Dismounting from the SUV  presupposes that it is stopped,
so C3 coerces the motion of the SUV in line 72 to  a ``stopped state.''
(Compare secret service agents dismounting from the presidential limo during a motorcade.)
If two people have {\it left} the SUV, qua container, then the number of people known to be in the vehicle (at least four) is reduced by two.

What has happened is that the introduction of the {\tt\small\bf dismount} to the situation initiated a limited inference process to identify the location the people dismounted from. Integrating the {\tt\small\bf dismount} with the established {\tt\small\bf enter} or the SUV provides a ``people-containing'' location to the inferential search (``inside the SUV''). 
If there had not already been such a location in the current situation,
the search would not go any further, and just posit that the location exists and wait for more information to come in, just as with our initial example of the Iraqi girl.


This example has illustrated our claim that language understanding is
an incremental process where every level of analysis is carried out simultaneously. 
We have shown how 
partial interpretations impose constraints on how they can be completed.
We have demonstrated the immediate effect of the implicatures 
conveyed by lexical packets 
(e.g.~every physical object has a location) 
in creating expectations as they are incorporated
into the situation, and leading to constrained searches of the
content of the situation organized by automatically-maintained indexical variables. 

\vspace{-3mm}

\section{Related Work}

We believe we have adopted a genuinely new perspective on 
deep natural language understanding. 
Others have worked on the same problems of course. 
Here we look at alternative approaches to 
gap inference, 
the use of frames, 
and parsing.

\paragraph{Gaps} To the best of our knowledge, the first person to describe ``gap
filling'' was Clark (1975), who  called this inferential
process \emph{bridging}.
\nocite{clark1975bridging}
He described it as ``the construction of implicatures'' in order to ``bridge
the gap from what [the listener] knows to what is intended''
(p.~170). 
In a logical framework, the process of adding implicatures in order
to make sense of a text is usually treated as a form of
\emph{abduction}. This sort of defeasible reasoning has been studied
at length by Hobbs (1993), who views language understanding as
finding the least-cost proof of the text's logical form.
%
Asher and Lascarides (1998) \nocite{asher1998bridging}
take a similar abductive approach to bridging inferences. 
They differ from Hobbs in using discourse cues and rhetorical relations to trigger their search
for suitable implicatures, and by running their search inside a 
representation of the text in their version of Discourse Representation
Theory.

We can also be said to be using abduction in that we add implicatures
to our minimal model of the situation
to bridge the gaps in the text (``the men had been in the SUV''). However
we do not follow Hobbs and formulate this as search over propositions and axioms by a
theorem prover. 
Instead, our approach is similar to Asher and Lascarides' use of DRT to
constrain where to look for implicatures. We use the structure
of our situation model to provide similar constraint. 
Moreover, we take the psycholinguistic evidence seriously and use an
architecture that anticipates inferences as latent variables 
that are deployed when a gap in the
text triggers them.

\paragraph{Frames}
In the 1970s, there was work on frame-based language understanding, but
either it formulated the problem in ways that could not be extended, 
such as the anticipated questions approach of Charniak (1975), or it made theoretical
assumptions that have since been rejected as psychologically
unrealistic and unnecessary: separating
syntactic, semantic and pragmatic analysis into cascaded independent
modules. 
\nocite{charniak1975organization}

During the 1980s and into the 1990s, frames devolved into just a way
to talk about a database record of related fields that
served as a template for the output of topic-specific information extraction
systems.
Over time, the semantics of these structures was clarified and the
result is today's description logics and the Web Ontology Language OWL \cite{horrocks2005owl}.
The original conception of frames as a way to manage perspectives and
provide defaults was lost. 

The Berkeley FrameNet Project is a curated effort to define the
meaning of concepts \cite{baker1998berkeley}.\footnote{
The FrameNet Project's use of the term ``frame'' derives from 
the linguistic notion of a ``case frame.''
}
It uses frames as hierarchically organized containers of
relationships, usually stated in terms of the standard Filmore case
relations (agent, patient, manner, etc.). 
FrameNet is a lexicalized ontology that we can draw on in our
research, but it is not suitable as a source for schemas to organize a situation.

\paragraph{Parsing}
Virtually all approaches to parsing today rely on training or
extending a probabalistic model and searching for the most likely
analysis given the features and corpus their models were developed on. There
is a body of recent work on what that community calls ``semantic
parsing'' 
\cite{kwiatkowski2011lexical}. However, they construe this as a
problem of recovering a sentence's logical form given matched pairs of
short texts and logical expressions as a training set. This is quite
different from our research on understanding a text in depth in order
to apply implicatures, establish predictive affordances, and
instantiate a model of the larger situation  a sentence is part of.

There are three other efforts that are engaged in the same kind of
high-precision, in-depth, linguistically principled language understanding work as we are. 
We all share a preference for rule-driven, largely deterministic
analysis based on a lexicalized conventional grammar. We all see the problem as
identifying the content of a text for some other program to use for
reasoning.
Clark and Harrison (2009) \nocite{clark2009large}
use a version of GPSG and have facilities for recognizing entailments
and other pragmatic phenomena. Much of their recent work is aimed at
adding to and querying a massive knowledge store
through a highly structured interface \cite{halo-2010}.
Allen's research group (2007) \nocite{allen2007deep}
focuses on the problems that occur in task-oriented dialogue. His group
uses a grammar based on a combination of GPSG and HPSG that is mapped
to a logical form that is grounded in a large general
ontology; task-specific representations are created by mapping from
that ontology. 
The LinGO group at Stanford and U. Washington (2010) has an extensive
semi-deterministic HPSG parser.  \nocite{bender2010grammar}
The output of their expressions is a set of formulas represented in
minimal recursion semantics \cite{copestake2005minimal}
that are comparable to those used by Hobbs (1993)
\nocite{hobbs1993interpretation} 
for abductive reasoning.

\vspace{-3mm}

\section{Future Work}



In this paper, we have presented a computational architecture
for a novel way to encode
and exploit the knowledge and inferences that make up a word's
meaning. Utterance meaning, we argue, involves the construction of ``cognitive simulations''
by the listener, of the situation being described. 
On this view, lexical  knowledge is composed of packets of frame-like structures, encoded as typing specifications, event and participant structures, and  {\it qualia} structure \cite{pustejovskyMIT1995}. In addition to this   enriched array of lexical semantic information, we introduced the notion of {\it habitats} \cite{pustejovskyGL2013}, a   data structure  that provides the conceptual underpinning for constructing the simulations compositionally. 
 This information is deployed by the processing mechanisms of Sparser \cite{mcdonald1996}, creating a dynamic interpretation of the event as it unfolds in the model.


%
Clearly, there is much  to be fleshed out, and it is difficult to
evaluate our proposal without more elaborate and extensive
modeling. One of the most promising and challenging aspects of this proposal is the exploitation of habitats in constructing a simulation. But questions remain, including the following: (i) how are habitats systematically constructed or related to the qualia structure associated with objects and events? (ii) what are the specific mechanisms of habitat composition, giving rise to minimal simulations that are cognitively plausible?
We are currently exploring these issues as they impact our design decisions for an efficient, robust, and incremental semantic interpreter.  
We believe that   the outline presented here suggests a specific way in
which people integrate and deploy their linguistic and general knowledge jointly to
understand discourse.

We originally intended to extend this small model 
to all five days of the Empire Challenge chat corpus. However, we discovered
that the  inferential gap illustrated by this example is unique; the rest of
the corpus can be understood with just a literal analysis. 
Consequently, we are shifting our future work to our original choice of
topic,  the inference-rich domain of following route directions in hiking guides. This
will let us develop vivid minimal simulation models and apply our
extensive background in spatial and temporal ontologies.

We welcome those who think that there is
merit in our goal  -- 
to understand how people can use their knowledge for language 
as quickly and effortlessly as they walk or breathe --
to engage in an extended conversation about how this is possible.

\vspace{-4mm}

 
\begin{acknowledgements} 

\vspace{-2mm}

\noindent
This work was supported in part by the Office of Naval Research under Grant N00014-13-1-0228. 
Any opinions, findings, and conclusions or recommendations expressed in this material are those of the authors and do not necessarily reflect the views of the Office of Naval Research.

\end{acknowledgements} 

\vspace{-0.25in}

\vspace{-4mm}

{\parindent -10pt\leftskip 10pt\noindent
\bibliographystyle{cogsysapa}
\bibliography{nl,kr,psych,ddm}

}


\end{document}

%% file: cup-macros.tex
{\obeyspaces\gdef {\ }}
\global\newbox\codebox
\global\newbox\savedcodebox
\gdef\sverbatim{\bgroup\def\endsverbatim{\egroup\egroup\egroup\mbox{\box\codebo\
x}}\def\savecode{\egroup\egroup\egroup\global\setbox\savedcodebox\copy\codebox}\
\def\par{\egroup\vspace{-0.3em}\hbox\bgroup}\tt\obeylines\obeyspaces\global\set\
box\codebox\vbox\bgroup\hbox\bgroup}

\newenvironment{enumerate*}%
  {\begin{enumerate}%
    \setlength{\itemsep}{0pt}%
    \setlength{\parskip}{0pt}}%
  {\end{enumerate}}

\newenvironment{itemize*}%
  {\begin{itemize}%
    \setlength{\itemsep}{0pt}%
    \setlength{\parskip}{0pt}}%
  {\end{itemize}}